# Language Lexicons for Hindi-English Multilingual Text Processing

**Mohd Zeeshan Ansari**[1*]
**Tanvir Ahmad**[1]
**Noaima Bari**[2]
*Department of Computer Engineering, Jamia Millia Islamia, India.*
*Department of Electrical Engineering, Jamia Millia Islamia, India.*

*\*mzansari@jmi.ac.in*

**Abstract**

Language Identification in textual documents is the process of automatically detecting the language contained in a document based on its content. The present Language Identification techniques presume that a document contains text in one of the fixed set of languages, however, this presumption is incorrect when dealing with multilingual document which includes content in more than one possible language. Due to the unavailability of large standard corpora for Hindi-English mixed lingual language processing tasks we propose the language lexicons, a novel kind of lexical database that supports several multilingual language processing tasks. These lexicons are built by learning classifiers over transliterated Hindi and English vocabulary. The designed lexicons possess richer quantitative characteristic than its primary source of collection which is revealed using the visualization techniques.

## 1. Introduction

Language identification (LI) is applicable to all forms of language, comprising spoken, sign, and handwritten and pertinent to any form of information storage that use language. The capacity to reliably recognize the language expressed in a document is an assisting technology which improves information accessibility and has a broad range of applications. Text processing approaches created in natural language processing and information extraction presume that the language of input text is known, and many approaches presume that all documents belong to the same language. Automatic language identification is used to ensure that only text in appropriate languages are submitted to further processing when applying text processing methods to real-world data. LI is required for document collections when the languages of documents are unknown a priori, such as data scraped from the internet, in order to prepare the multilingual index of documents collection by the language in which they were authored. The identification of a document's language for routing to an appropriate translation is another use of LI that precedes computational approaches. LI has gained prominence as a result of the emergence of Machine Translation, which need the source language of the text to be determined first. LI significantly helps with documentation and usage of low-resource languages.

Language Identification is an old problem in Natural Language Processing (NLP) that has been widely studied both in the speech (House and Neuburg, 1977) and text (Cavnar and Trenkle, 1994) domain [1,2]. The task of Language Identification in mixed lingual documents is defined as the automatic detection of language(s) present in the document at sentence, phrase or word level based on the content of the document [3]. Many existing LID techniques presume a document to contain text only in a single language from a given set of languages. However, this presumption stands untrue while dealing with multilingual documents especially found on the web, that contain text from more than one language from the candidate set. Further, most of the NLP systems assume input data to be monolingual in nature and by the inclusion of data from foreign languages in such systems, noise is introduced and performance degrades (Baldwin, Cook & Lui, 2012) [4]. One of the major challenges in Language Identification in multilingual documents stems from a shortage of labelled multilingual text for training the LI models. Standard corpora of multilingual documents are rarely available, whereas corpora of monolingual documents are readily available even for a reasonably large number of languages (Lui and Baldwin, 2011) [5]. Lexical databases such as Wordnet and BabelNet are collections of entries, each of which contains hierarchical text that provides information on a particular concept [19,20]. The bulk of such lexical databases, including conventional dictionaries, are relational in form and entirely textual in content. Their organizational structure does not represent to the quantitative nature of words. To address this issue, we created a Hindi-English dataset, with minimal human intervention, by integrating different monolingual language corpora, subsequently, produced the lexicons with language strength associated with them.

The present language models are capable of capturing sufficient semantic information, however, they fail to differentiate the language present in the text. Moreover, morphological characteristics are explicitly extracted as features from the text for language identification. It is therefore important to design the models that can condense the linguistic strength of each word and represent them with interpretable lexicons. The language lexicons presented in this work augment the language models with additional compact information which helps to discriminate between texts in separate languages. We develop domain-independent Hindi-English language lexicons utilizing monolingual corpora that enhance the language-aware learning models capable of performing a variety of language processing tasks such as information extraction, sentiment analysis, etc.

The lexical structures we propose are not only focused on modelling a few particular linguistic characters, but rather modelled from a broader view of the lexicon as a key component of language strength. Secondly, they feature a very straightforward, flat structure that does not impose any ordered or hierarchical structure on the vocabulary. Thirdly, they deal with the Out of Vocabulary problem due to its inherently coupled character level methodological design.

## 2. Related Work

Many existing research works on document-level language identification consider only mono-lingual documents. However, the task of LI is far from solved, particularly, when dealing with multilingual documents, short texts and informal styles such as those found in the real world and social media platforms, or while working with language pairs which are closely related. (Baldwin and Lui, 2010, Gella et al.,2014; Wang et al., 2015) [6-8].

Many earlier works in the text domain have utilized word or character n-gram features followed by linear classifiers (Ansari et al) [9]. A bleak picture depicting support for low-resource languages in automatic language identification has been painted by Hughes et al. (2006) [3]. King and Abney (2013) propose HMM and CRFs for labelling words in a multilingual corpus and frame LI as a sequence labelling problem [10]. Language identification in multilingual documents is performed using a generative mixture model combined with a document representation. The model learns a language identifier for multilingual documents from monolingual training data which is more abundant as compared to labelled multilingual textual data [11].

Word-level language identification was largely addressed using supervised techniques. For example, King and Abney (2013) show that the problem can be framed as a sequence labelling problem and that using hidden Markov Models (HMMs) and Conditional Random Fields (CRFs) [10]. The problem can be trained to perform reasonably well at labelling words in multilingual texts starting with monolingual data. In the first shared task on LID on code-switched data in 2014, system designs varied from rule-based systems to those that used word embeddings, enhanced Markov Models, and CRF autoencoders (Solorio et al., 2014) [12]. While most teams focused on multilingual LI systems for the shared task, there are approaches that specifically deal with classification on bilingual code-switched texts. For example, Jhamtani et al. (2014) built a system that uses several heuristic features, including a special edit distance between Hindi and English that fits their use case for "Hinglish" texts [13].

Unsupervised techniques to language identification at the word level have not proven as popular [10]. Rabinovich and Wintner (2015) used a cluster-and-label strategy to discover unsupervised "Translationese" for machine translation, but only for text passages of 2000 tokens [15]. The cluster labeling methodology is another feature that distinguishes their technique from ours. Their automated labeling method creates "representative" language models for the class labels, which are subsequently assigned to the unsupervised clusters by comparing them to the clusters' empirical distributions. We resorted to manual labeling since their methodology is not appropriate in our instance because tagging clusters for our purpose involves very little work.

Some researchers have utilized models based on artificial neural networks in addition to the usual machine learning methodology. For the SPA-ENG and Nepali-English datasets from the First Shared Task on Language Identification in Code-Switched Data, Chang and Lin (2014) employ an RNN architecture with pre-trained word2vec embeddings [16]. For the SPAENG and MSA-DA datasets from the Second Shared Task on Language Identification in Code-Switched Data, Samih et al. (2016) built an LSTM-based neural network architecture. Their model blends pretrainedword2vec embeddings with word and character representations [17].

Using a Hindi–English code-mixed speech corpus from student interviews, Dey and Fung (2014) examined the grammatical contexts and motives. Many researchers have looked at the detection of code-mixing in text [23]. The prediction of the places in Spanish–English phrases when the orator transitions between the languages was started by Solorio and Liu (2008) [24]. [25–27] investigates code-mixing in brief messages and information retrieval queries. Nguyen and Dogruoz (2013) utilized various linguistic models, dictionaries, and probabilistic models such as Conditional Random Fields and Logistic Regression to experiment on Turkish and Dutch forum data [28].

| Table 1 | | | | | |
|---|---|---|---|---|---|
| Data statistics for Hindi-English lexicon dataset | | | | | |
| | #counts | %age | max word length | average word length | example words |
| **English** | 25640 | 70.38 | 20 | 7.63 | *public, deception, great, synchronous, convinced, dramatically* |
| **Hindi** | 10789 | 29.62 | 15 | 6.81 | *gulaam, ,khiladii, paayal, samudr, himmatawala, mangaladaata* |

# 3. Data Collection

### a. Obtaining the lexicons and tagset

For the preparation of the Hindi-English dataset two separate rich language sources are utilized. Hindi words were taken from the Hindi Transliteration Dataset given by [18] consisting of 30696 Hindi words written in their transliterated form in the roman script. The dataset was manually filtered by 2 annotators having Hindi as their native language, in order to remove incorrectly present words such as "everybody", "sing", "something" etc. not belonging to the Hindi language, hence, leading to a disagreement score of 0.4% over the unknown words. Further, after the removal of duplicate words, trailing spaces and new lines, 25640 unique Hindi words were obtained that were annotated as Hindi (Hi) thus forming the Hindi words dataset.

The English words were extracted from the frequent word list from BNC corpus consisting of 13000 words. A number of hyphenated words such as "self-confidence", "fund-raiser" present in the list were broken down and added as separated words as "self", "confidence", "fund" "raiser". Further, duplicate words were removed and the final list of English words consisted of 10789 words annotated with the label English (En).

Finally, both the English and Hindi datasets were combined into one single one consisting of 36429 words appropriately annotated into 2 classes – Hindi (hi) and English (en) and randomly shuffled. The final dataset contained some amount of class imbalance with English and Hindi words comprising about 29.62% and 70.38% of the dataset respectively. The Table 1 summarizes the distribution of Hindi-English language tags in our dataset and word lengths of the prepared dataset. We see that there is indeed a high fraction of Hindi tags.

# 4. Methodology

The Lexicon dataset is used for building and evaluating an automatic Language Identification model. We tokenize the words at character level and train a BiLSTM classifier with softmax classification. The softmax output for Hindi Tags is considered as the score for language strength. We also present a set of n-gram features using which the logistic regression learns to predict the language tag of a token and, subsequently, generates the second score.

### a. Learning the classifiers

The prepared dataset is used to learn two classifiers BiLSTM and Logistic Regression respectively.

(i) **LSTM**: The recurrent neural networks (RNNs), are a kind of neural network that operates with sequential input. They accept a series of vectors as input and output a new sequence that provides information about the sequence at each step in the input. Although RNNs are capable of learning lengthy dependencies in principle, they do not do so in reality and are biased towards the most recent inputs in the sequence (Bengio et al., 1994) [21]. Long Short-Term Memory Networks (LSTMs) have been found to capture long-range dependencies and have been built to overcome this problem by integrating a memory-cell. They accomplish so by controlling the quantity of the input to deliver to the memory cell, as well as the percentage of the previous state to forget, utilizing several regulatory gates (Hochreiter and Schmidhuber, 1997) [22]. The LSTM computes a representation of the left context of a character sequence having n words, each represented as a d-dimensional vector. Developing a representation of the appropriate directions is accomplished by reading the same sequence in reverse using a second LSTM. The former will be referred to as the forward LSTM, while the latter will be referred to as the backward LSTM. These are two separate networks, each with its own unique set of parameters. A bidirectional LSTM is a pair of forward and backward LSTMs. A word is represented in this model by concatenating its left and right context representations. These representations effectively incorporate a contextual representation of a word, which is useful for a variety of applications.

(ii) **Logistic Regression**: Logistic regression is a popular method for examining and defining a connection between a binary response variable and a collection of input variable. Logistic regression is linear classifier that fits data to the logistic function and predicts the probability of an event occurring. In order to train the classifier, we utilize the n-grams with n=1-5 as features.

b. **Language Score**

The language strength of the lexicons are generated according to the scoring methods based on the LSTM and Logistic Regression classifiers respectively.

(i) **Score 1**: It is calculated according to the following softmax function

$$score(x) = \frac{e^{O_j}}{\sum_k e^{O_k}} \tag{1}$$

where $O_j = BiLSTM(X)$ is the output from the last layer of the BiLSTM architecture that corresponds to the Hindi Language tag.

(ii) **Score 2**: It is calculated according to the following logistic function

$$score(x) = \frac{e^{\theta_0 + \theta_1 X_1 + \ldots + \theta_p X_p}}{1 + e^{\theta_0 + \theta_1 X_1 + \ldots + \theta_p X_p}} \tag{2}$$

where $X_1 + X_2 + \ldots + X_p$ is the input feature set and $\theta_1 + \theta_2 + \ldots + \theta_p$ are the parameters learned from the dataset.

The score1 and score2 calculated from (1) and (2) represent the Hindi language strength of the words present in the lexicons. The score denotes a value between 0 and 1. The score of a word close to 1 is considered as a lexicon with very high Hindi language strength and vice-versa.

| Table 2 Performance of classifiers on lexicon dataset | | | | | | | |
|---|---|---|---|---|---|---|---|
| | | Precision | | Recall | | F-Score | |
| Model | | LSTM | LR | LSTM | LR | LSTM | LR |
| English | 2697 | 92.61 | 95.74 | 88.25 | 91.77 | 90.37 | 93.71 |
| Hindi | 6411 | 95.15 | 96.60 | 97.04 | 98.28 | 96.08 | 97.43 |
| Weighted Avg | 9108 | 94.40 | 96.34 | 94.43 | 96.35 | 94.39 | 96.33 |

# 5. Experiments and Results

We compare the performance of our models and illustrate the effect of obtained lexicons by showing the boxplots of language strength scores and by visualizing their tags in a two dimensional scatter plot. For the experiments, we hold out 10% of the data as a test set. We perform parameter tuning for 90% and report the performance of our models on the test set in Table 2. The reported results are the best among the several runs of models overs different random samples of training dataset.

a. **Classifier Performance**

The results of the investigation in Table 2 reveal that when the models are trained with 27,321 training instances, the overall performance of the classifiers with 9,108 test samples is significant. When the precision, recall, and F-Score of each class are compared, it is discovered that the predictions of the Hindi class are more accurate than those of the English class, though with a small but significant difference. However, the performance of the classifiers demonstrates that the Logistic Regression model outperforms the BiLSTM model in terms of all the measures. Although the Logistic regression produces the highest F-Score of 96.33, its recall is pretty similar with a difference of 1.24 percent only with BiLSTM.

b. **Lexicon Analysis**

We analyze the suggested lexicons language strength by examining the scatter plot and box plot of the scores obtained from (1) and (2) as presented in Fig. 1. The scatter plot displays the score1 on x-axis and score2 on y-axis with the blue data points representing English class and orange data points representing the Hindi Class. It demonstrates that the lexicons are clearly separable with small quantity of outliers, hence, when the other Hindi-English Multilingual datasets are supplemented with these lexicons, the models should perform better, which is consistent with our proposed concept of language lexicons.

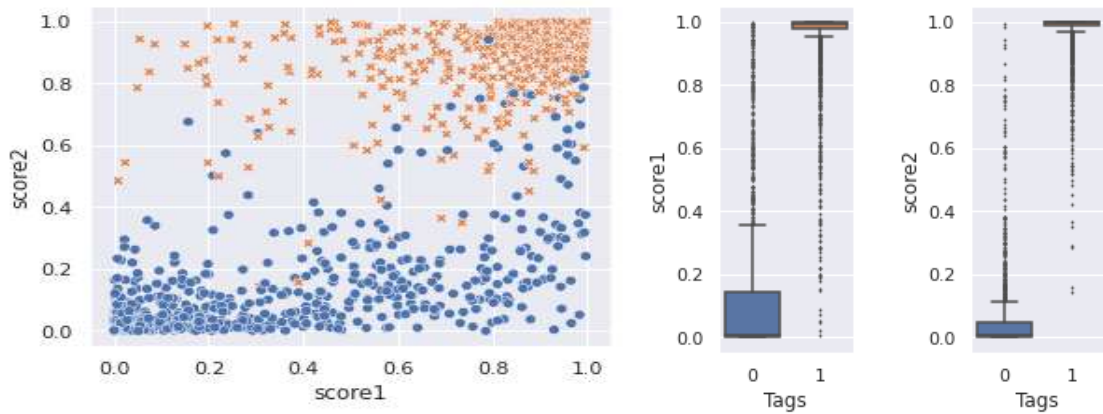

**Fig. 1**

The box plots of Fig. 1 reveal that the scores of Hindi words (denoted by Tag 0) have very high language strength being majority of scores very close to 1. On the other hand, the English words have very low Hindi language strength with majority scores less than 0.2. Table 3 shows some sample lexicons generated the proposed methodology which includes some outliers also.

# 6. Conclusion

We have created a considerable size of language lexicons that may be regarded comprehensive in terms of the number of entries and quantifiable language strength to indicate the extent of linguistic information it stores. The language lexicons created by utilizing the complementary English and Hindi Roman vocabulary. The lexicons show the Hindi linguistic power of each word in a two-dimensional space. The study of lexicons acquired shows that they have acquired language characteristics such that it has high values for Hindi words and low values for English words. Additionally, the proposed models may be used to assess the language strength of new words and may be integrated with any kind of multilingual model.

Table 3
Sample Language Lexicons

| Word | Score1 | Score2 |
|---|---|---|
| abhilaasha | 0.9943395256996155 | 0.9999969538347858 |
| baharoon | 0.9996402263641357 | 0.9969923646323342 |
| khuski | 0.9981417655944824 | 0.9987074582327405 |
| gulaam | 0.9978258013725281 | 0.996956514189548 |
| literally | 0.9714275598526001 | 0.20443189077443946 |
| jurisdictional | 0.7277640104293823 | 0.17373738918922257 |
| reinforce | 0.7902541756629944 | 0.14437307870695257 |
| suspension | 0.9440618753433228 | 0.2587238402918156 |
| intazaam | 0.7656338810920715 | 0.30991121414455586 |
| bikhare | 0.8590598106384277 | 0.41401772007563414 |
| hamsafar | 0.7710220813751221 | 0.26723702194071386 |